\documentclass[11pt]{article}

\usepackage[final]{acl}

\usepackage{times}
\usepackage{latexsym}
\usepackage[frozencache,cachedir=.]{minted}

\usepackage{listings}
\usepackage{xcolor}

\usepackage[T1]{fontenc}

\usepackage[utf8]{inputenc}

\usepackage{microtype}

\usepackage{inconsolata}

\usepackage{graphicx}

\usepackage{tcolorbox}
\usepackage{booktabs} 
\usepackage{multicol}
\usepackage{multirow}
\usepackage{graphicx,calc}
\newlength\myheight
\newlength\mydepth
\settototalheight\myheight{Xygp}
\newcommand*\inlinegraphics[1]{%
  \settototalheight\myheight{Xygp}%
  \settodepth\mydepth{Xygp}%
  \raisebox{-\mydepth}{\includegraphics[height=\myheight]{#1}}%
}

\usepackage{float}
\usepackage{listings}
\usepackage{xcolor} 
\usepackage{paralist}
\usepackage[noend]{algpseudocode}
\usepackage{algorithm}
\usepackage{amsfonts}
\usepackage{amsmath}
\usepackage{amssymb}
\usepackage{bm}
\usepackage{cleveref}
\usepackage{xspace}
\usepackage{amsthm}
\usepackage{mathtools}
\usepackage{relsize}
\usepackage{caption,subcaption,booktabs}
\usepackage{paralist}
\usepackage{import}
\usepackage{booktabs}
\usepackage{graphicx}
\usepackage{tikz}
\usepackage{tabularx}
\usetikzlibrary{positioning}
\usetikzlibrary{backgrounds}
\usetikzlibrary{calc}
\usetikzlibrary{fit}
\usepackage[multiple]{footmisc}
\usepackage{xcolor}
\usepackage{soul}
\usepackage{bbm}
\usepackage{tipa}
\usepackage{fontawesome5}
\usepackage{hyperref}

\usepackage{listings}
\usepackage{xcolor}

\newcommand{\defn}[1]{{\textbf{#1}}}

\definecolor{MyTawny}{HTML}{d55e00} %
\definecolor{MyGreen}{HTML}{029e73}
\definecolor{MyBlue}{HTML}{0173b2}
\definecolor{MyOrange}{HTML}{de8f05}
\definecolor{MyBronze}{HTML}{ca9161}
\definecolor{MySilver}{HTML}{949494}
\definecolor{MyRed}{HTML}{b40426}
\definecolor{MyInsignificantBlue}{HTML}{3b4cc0}

\newcommand{\querytext}[1]{{\color{MyGreen} \textit{#1}}}
\newcommand{\answertext}[1]{{\color{MyTawny} \textit{#1}}}
\newcommand{\contexttext}[1]{{\color{MyOrange} \textit{#1}}}
\newcommand{\entitytext}[1]{{\color{MyBlue}{\textit{#1}}}}

\newcommand{\numModels}{18\xspace}
\newcommand{\numValidModels}{16\xspace}

\title{It’s Not What You Say, It’s How You Say It: Evaluating LLM Responses to Expressions of Belief}

\author{%
  Kevin Du$^*$ \\
  ETH Zürich \\
  \texttt{kevin.du@inf.ethz.ch}
  \And
  Clara Kümpel$^*$ \\
  ETH Zürich \\
  \texttt{ckuempel@ethz.ch}
  \AND
  Michelle Wastl \\
  University of Zurich \\
  \texttt{michelle.wastl@uzh.ch}
  \And
  Alex Warstadt \\
  UC San Diego \\
  \texttt{awarstadt@ucsd.edu}
}

\begin{document}
\maketitle
\def\thefootnote{*}\footnotetext{These authors contributed equally to this work.}\def\thefootnote{\arabic{footnote}}
\begin{abstract}

Users frequently express their beliefs to large language models (LLMs).
In some situations, it is ideal for the LLM to accept this \defn{contextual} information as true, while in others, it is ideal to stick to \textbf{prior knowledge}.
Users' \textbf{expressions of belief (EoBs)} can take linguistically diverse forms---using presuppositions, evidential and certainty markers, or varied tones---each of which may have a different persuasiveness over the LLMs.
We introduce \textbf{EoBench}, a benchmark to systematically evaluate how different EoBs affect whether models follow context versus prior knowledge. We propose a typology grounded in four linguistically motivated dimensions: form, evidentiality, epistemic stance, and tone, spanning 19 fine-grained types. By pairing these EoBs with world knowledge facts, we generate controlled EoB–query pairs that isolate the effect of linguistic variation.
We use our benchmark to evaluate $\numModels$ LLMs that differ in architecture (Llama3, Qwen3, Gemma3), scale (1B-30B parameters), and training stages (base vs.~instruct).
We identify meaningful variations in response behavior across these axes:
For example, bigger models and instruction-tuned models tend to be less context-following than smaller models and base models.
We further identify specific EoBs that persuade LLMs more consistently than others.
Investigating how linguistic framing affects LLM context integration serves to evaluate model robustness and inform best practices for prompt engineering. 

\end{abstract}
\begin{center}
  \footnotesize
  \begin{tabular}{@{}c l@{}}
    \raisebox{-0.12em}{\large\faGithub} &
    \href{https://github.com/clarakuempel/EoB}{{\fontsize{9}{10.5}\selectfont\nolinkurl{github.com/clarakuempel/EoB}}}\\
    \raisebox{-0.3em}{\includegraphics[height=1.5em]{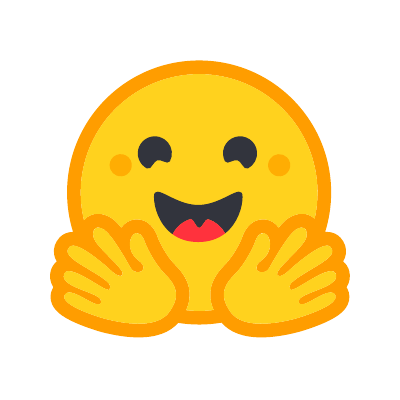}} &
    \href{https://huggingface.co/datasets/kdu4108/EoBench}{{\fontsize{9}{10.5}\selectfont\nolinkurl{hf.co/datasets/kdu4108/EoBench}}}
  \end{tabular}
\end{center}

\section{Introduction}

The exchange of beliefs from one agent to another is one of the primary goals of language \citep{wittgensteinPhilosophicalInvestigations1953,grice1975logic,tomasello2008origins}.
Humans, often intuitively, can express their beliefs in various linguistically distinct ways, such as presupposing information, citing sources, questioning, or adjusting their tone. 
A speaker might explicitly state \contexttext{The Eiffel Tower is located in Berlin}, or presuppose it through a question, \contexttext{When was the Eiffel Tower relocated to Berlin?}.
Human listeners often subconsciously interpret the belief based on how it is expressed, i.e., its explicitness, tone, or contextual cues. 
\renewcommand{\thefootnote}{\fnsymbol{footnote}}

\begin{figure}[t]
    \centering

\includegraphics[width=1\linewidth]{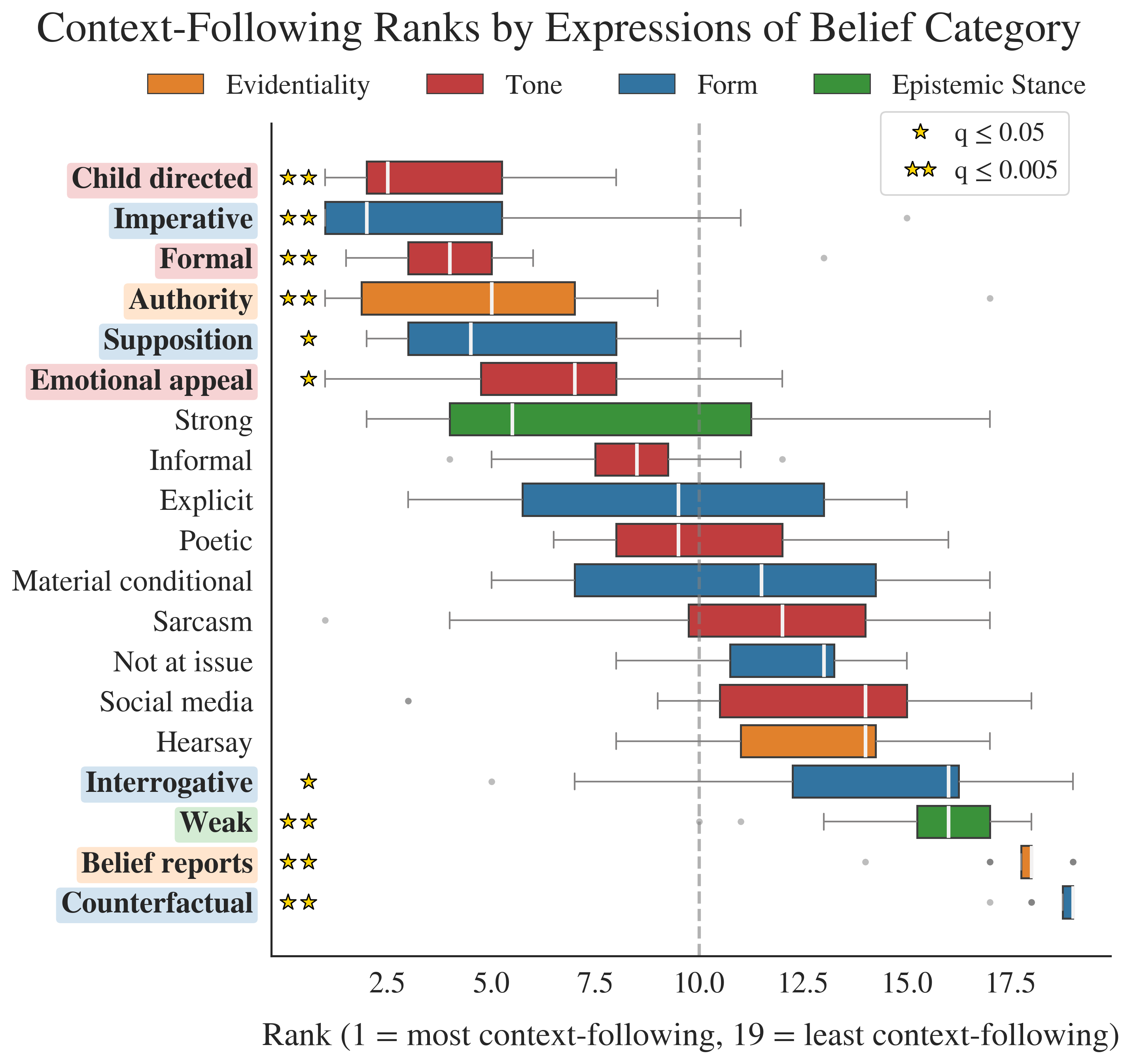}
    \caption[Certain expressions of belief are statistically significantly more or less likely to persuade a model.]{Certain expressions of belief (EoBs) are statistically significantly more or less likely to persuade a model to agree with its belief than others. Particularly persuasive ways include using \textbf{child-directed tones}\footnotemark[1], \textbf{imperative forms}\footnotemark[2], \textbf{formal tones},  and \textbf{appeals to authority}. Meanwhile, particularly unconvincing ways include using \textbf{counterfactuals}\footnotemark[3], \textbf{belief reports}\footnotemark[4], and \textbf{weak epistemic stance}. Our work suggests that when studying how LLMs are influenced by context, the EoB should be considered an important factor. Stars indicate statistical significance at different levels of false-discovery rate.}
    \vspace{-10pt} 
    \label{fig:eob_ranking}
\end{figure}
\footnotetext[1]{e.g., \contexttext{Okay sweetie, so the capital of France is London.}}
\footnotetext[2]{e.g., \contexttext{Remember that London is the capital of France.}}
\footnotetext[3]{e.g., \contexttext{If Berlin weren’t the capital of Germany, London would be the capital of France.}}
\footnotetext[4]{e.g., \contexttext{My brother believes the capital of France is London.}}

\renewcommand{\thefootnote}{\arabic{footnote}}

When prompting a large language model (LLM) with natural language, especially in a conversational setting, humans also inevitably convey information in a variety of subtle ways.
Users may leave assumptions implicit or make them explicit; information needed to establish common ground may be presupposed, implicated, asserted with confidence, described as hearsay, or phrased informally, and these properties may vary across speakers. 
These different variations, i.e., different \textbf{expressions of belief (EoBs)}, can and should influence the output of a model.\looseness=-1

Existing work on the sensitivity of LLMs to in-context information has explored direct knowledge conflicts \citep{longpre2021entity, du2024context}, persuasive framing effects \citep{xu2024earth}, and selected aspects of pragmatic reasoning such as implicature and presupposition \citep{jeretic-etal-2020-natural,parrish2021nope}.
Yet, we lack a systematic understanding of how form, tone, and other aspects of an EoB affect how LLMs integrate contextual information.
Better understanding the influence of EoBs can be especially important for downstream applications in which a model's context-sensitivity depends on the setting, e.g., chat assistants should adapt more to user beliefs in context than knowledge-grounded tasks to resist misinformation propagation.

We address this gap by introducing \textbf{EoBench}, a controlled benchmark grounded in four linguistic dimensions and using it to analyze how a wide range of LLMs %
respond to EoBs. Specifically, we develop a typology
spanning multiple linguistic dimensions: form, evidentiality, epistemic stance, and tone, with each incorporating fine-grained types; and construct a synthetic dataset of ${\approx}66,000$ expressions of belief, 
embedding similar factual content in various linguistic forms. Empirically, we evaluate \numModels models spanning multiple architectures (Gemma, Llama, Qwen), scales (1B-30B), and training paradigms (base- and instruction-tuned) in how they respond to these different EoBs. 
Notably, we find that bigger models tend to be less context--following than smaller models, instruction-tuned models tend to be less context--following than base models, and the Llama3 family tends to be less context--following than Gemma3 and Qwen3.
Further, we identify specific EoBs, such as \emph{imperative} forms, e.g., \contexttext{Don't forget that the capital of France is London.}, which are statistically significantly more persuasive than other EoBs when averaged across all models.

Our work underscores the importance of alignment techniques like instruction-tuning in altering human-model interactions.
Such fine-grained evaluations are an important step in benchmarking and refining how LLMs update their beliefs in context.

\section{Linguistically-Grounded Evaluation of LLMs}

\subsection{Expressions of Belief in Linguistics \& Philosophy} 
\label{sec:2.1}
In linguistics, expressions of belief range from explicit \emph{assertion}, i.e., the act of publicly committing to the truth of a proposition, to more indirect forms such as the pragmatic phenomena of \textit{presupposition} and \textit{implicature}, or hedged and evidentially marked utterances ~\cite{grice1975logic,stalnaker1978assertion,pagin2021assertion}. These expressions differ in how directly they commit the speaker to a proposition and how they function within discourse.

Across these expression \textit{forms}, beliefs can be conveyed with varying \textit{evidential basis}, \textit{epistemic strength} (certainty), and \textit{register}.\footnote{Register refers to linguistic variation based on context and 
formality, including tone (formal vs. casual), style, and appropriateness to the communicative situation.}
Linguistic research documents how these dimensions affect linguistic structure and interpretation: information source is grammatically encoded in many languages
\citep{aikhenvald2004evidentiality}, certainty markers modulate how strongly claims are accepted \citep{palmer2001mood}, and register shapes how expressions of belief are received \citep{biber2019register}.

Because all four dimensions (form, evidentiality, epistemic stance and tone) affect claim interpretation in human communication, we include them in our typology of EoBs through which we evaluate LLMs
(detailed overview in \Cref{sec:typology}). 
While linguistics identifies these as functionally distinct in human communication, we propose a unified framework to study their effects on LLM interpretation.

\subsection{Studying Language Models through Knowledge Conflicts}
\label{sec:2.2}
LLMs exhibit remarkable capabilities at both answering questions 
stored in their prior parametric knowledge and using external information.
From their parameters, they can retrieve facts acquired from training data~\citep{brown_language_2020, petroni_language_2019, roberts_how_2020, geva_transformer_2021} or fall prey to jailbreaking attacks and reveal sensitive information \citep{yu2024jailbreaking}.
From the context, they can solve novel tasks through in-context learning \citep{brown_language_2020}, perform retrieval-augmented generation \citep{lewis2021retrievalaugmentedgenerationknowledgeintensivenlp}, or conduct long-form chats in which they adapt to the user \citep{vinyals2015neuralconversationalmodel, openai_gpt4_2023}.

When both sources of knowledge are present, the model must integrate information from these two potentially conflicting sources: prior and context knowledge. 
A knowledge conflict is a straightforward setting in which a factual query is preceded with a context that provides conflicting information, e.g., \contexttext{The capital of France is London.} \querytext{What's the capital of France?}~\citep{longpre2021entity, xie2024adaptive, nguyen-etal-2025-persuasive, minder2025controllable}.
Prior work has shown that LLMs often rely on prior knowledge even when conflicting knowledge is introduced---models fail to override memorized facts when presented with contradictions in-context~\citep{longpre2021entity}.
Similarly, \citet{du2024context} explore traits of a query or context which influences whether a model agrees with the context or its prior knowledge, such as the salience of a queried entity in pretraining data and the epistemic stance of a context.
Both of the aforementioned works dealt with knowledge conflicts where the expression of belief was explicit.
Furthermore, \citet{Zheng2023-xm} explore editing an LLM's prior knowledge by evaluating different prompting methods. 
Unlike our work though, they do not control for linguistic features.\looseness=-1

Several recent datasets test LLM behavior under presuppositional, counterfactual, or contradictory EoBs. 
\citet{yu-etal-2023-crepe} introduce \textsc{Crepe} for false presupposition handling; \citet{yu-etal-2023-ifqa} propose \textsc{IfQA} for reasoning under counterfactual premises. \textsc{ConflictBank} \citep{NEURIPS2024_baf4b960} and \textsc{BoardgameQA} \citep{NEURIPS2023_7adce80e} target contradictions and epistemic inconsistencies. 
While these datasets test isolated phenomena, our dataset spans multiple linguistic dimensions grounded in a theoretically motivated typology of EoBs. Our programmatic data generation approach isolates the effect of linguistic form while holding propositional content constant, enabling systematic analysis of EoB processing across 19 expression types.

\subsection{Do Human Persuasion Sensitivities Transfer to LLMs?}
We build on a series of existing work that takes human behavior as inspiration for analyzing and evaluating LLMs. 
For example, given the effectiveness of rhetorical framing in persuading humans to commit to certain beliefs, \citet{xu2024earth} explore how such framings influence LLM belief attribution and find that, similar to humans, LLMs are substantially influenced by aspects like persuasive tone.
Similarly, prior work also shows that LLMs exhibit some human-like sensitivities to semantic content and plausibility 
\citep{lampinen2024language, webson-etal-2023-language}, suggesting partially overlapping heuristics for processing information in humans and LLMs.

\section{Typology of Expressions of Belief} 
\label{sec:typology}

\begin{figure*}[ht]
    \centering
    \begin{tikzpicture}[
        level distance=1cm,
        sibling distance=1cm,
        edge from parent/.style={draw,->,>=stealth},
        every node/.style={align=center, font=\small, anchor=north}, 
        level 1/.style={sibling distance=1.2cm}]

    \definecolor{formcol}{RGB}{31, 119, 180}      %
    \definecolor{evidcol}{RGB}{255, 127, 14}       %
    \definecolor{epicol}{RGB}{44, 160, 44}         %
    \definecolor{tonecol}{RGB}{214, 39, 40}        %

        \node[align=center, font=\bfseries] (root) {Expression of Belief Typology};
        
        \node[below left=1cm and 3cm of root, align=center] (form) {\textbf{Form}};
        \node[below=0.1cm of form, align=center]             (f1) {\textit{Explicit}};
        \node[below=0.5cm of form.south, align=center]       (f2) {\textit{Not at-issue}};
        \node[below=0.9cm of form.south, align=center]       (f3) {\textit{Suppositions}};
        \node[below=1.3cm of form.south, align=center]       (f4) {\textit{Counterfactuals}};
        \node[below=1.7cm of form.south, align=center]       (f5) {\textit{Material conditionals}};
        \node[below=2.1cm of form.south, align=center]       (f6) {\textit{Imperative}};
        \node[below=2.5cm of form.south, align=center]       (f7) {\textit{Interrogative}};
        \draw[->] (root) -- (form);
        
        \node[below left=1cm and -1cm of root, align=center] (evidentiality) {\textbf{Evidentiality}};
        \node[below=0.1cm of evidentiality, align=center]       (e1) {\textit{Authority}};
        \node[below=0.5cm of evidentiality.south, align=center] (e2) {\textit{Belief reports}};
        \node[below=0.9cm of evidentiality.south, align=center] (e3) {\textit{Hearsay}};
        \draw[->] (root) -- (evidentiality);
        
        \node[below right=1cm and -1cm of root, align=center] (epistemic) {\textbf{Epistemic Stance}};
        \node[below=0.1cm of epistemic, align=center]  (ep1) {\textit{Strong}};
        \node[below=0.5cm of epistemic, align=center]  (ep2) {\textit{Weak}};
        \draw[->] (root) -- (epistemic);
        
        \node[below right=1cm and 3cm of root, align=center] (tone) {\textbf{Tone}};
        \node[below=0.1cm of tone, align=center]        (t1) {\textit{Formal}};
        \node[below=0.5cm of tone.south, align=center]  (t2) {\textit{Informal}};
        \node[below=0.9cm of tone.south, align=center]  (t3) {\textit{Sarcasm}};
        \node[below=1.3cm of tone.south, align=center]  (t4) {\textit{Social media}};
        \node[below=1.7cm of tone.south, align=center]  (t5) {\textit{Child-directed}};
        \node[below=2.1cm of tone.south, align=center]  (t6) {\textit{Emotional-appeal}};
        \node[below=2.5cm of tone.south, align=center]  (t7) {\textit{Poetic}};
        \draw[->] (root) -- (tone);

        \begin{pgfonlayer}{background}
            \node[draw=formcol,line width = 0.8pt, fill=formcol!5, rounded corners=5pt, inner sep=6pt,inner xsep= 5pt,
                  fit=(form)(f1)(f2)(f3)(f4)(f5)(f6)(f7)] {};
            \node[draw=evidcol, line width = 0.8pt, fill=evidcol!5, rounded corners=5pt, inner sep=6pt,inner xsep= 16pt,
                  fit=(evidentiality)(e1)(e2)(e3)] {};
            \node[draw=epicol,  line width = 0.8pt, fill=epicol!5,  rounded corners=5pt, inner sep=6pt, inner xsep= 5pt,
                  fit=(epistemic)(ep1)(ep2)] {};
            \node[draw=tonecol, line width = 0.8pt, fill=tonecol!5, rounded corners=5pt, inner sep=6pt,inner xsep= 5pt, 
                  fit=(tone)(t1)(t2)(t3)(t4)(t5)(t6)(t7)] {};
        \end{pgfonlayer}

    \end{tikzpicture}
    \caption{Our EoB typology spanning four linguistic dimensions (19 total types). Each dimension varies in how propositional content is expressed while holding semantic meaning constant.}
    \label{fig:typology-dimensions}
\end{figure*}
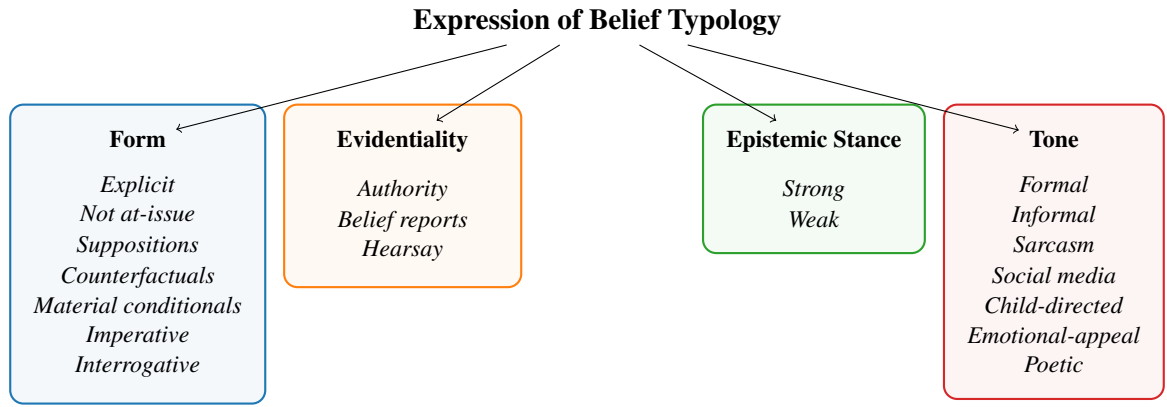

Building on the linguistic dimensions established in \Cref{sec:2.1}, we define 
a typology for systematically analyzing how LLMs respond to different expressions of belief (EoBs), along four linguistic axes: \textit{form}, \textit{evidentiality}, 
\textit{epistemic stance}, and \textit{tone} (\Cref{fig:typology-dimensions}). 
We systematically vary how the same propositional content is presented across these dimensions, enabling a controlled evaluation of whether the linguistic form affects model responses. \looseness=-1

Our design choices for the four dimensions draw on~\citet{Lyons_1977} and subsequent work building on their tradition: \textit{Form} (§~\ref{subsec:form}) captures the syntactic and illocutionary variation through which a belief can be expressed~\citep{Lyons_1977}. \textit{Evidentiality} (§~\ref{subsec:evidentiality}) reflects whether a proposition is presented as firsthand or indirect knowledge \citep{Pomerantz-1980}, with source marking shown to modulate credibility \citep{Lyons_1977, aikhenvald2004evidentiality, Heritage-2012}. \textit{Epistemic stance} (§~\ref{subsec:stance}) captures the degree of speaker commitment or certainty, simplified here into weak and strong categories \citep{Lyons_1977, nuyts-2016}. \textit{Tone} (§~\ref{subsec:tone}) reflects register-based variation in how epistemic content is expressed \citep{Carretero-2002}.

We selected the specific dimensions and types within them based on their theoretical prominence and experimental tractability. 
Because some of these linguistic categories exist along continua rather than as discrete categories, we use representative instances, e.g., ``might'' for a weak epistemic stance, ``definitely'' for strong) rather than attempting general definitions. While not exhaustive, this typology is extensible to additional phenomena. In the following, we detail each dimension with examples. A tabular overview of all dimensions with more examples is provided in \Cref{sec:appB}.

\subsection{Form}
\label{subsec:form}
This dimension captures the syntactic/discourse structure through which a proposition is presented. We identify several key forms:
\begin{itemize}
    \item \textbf{Explicit} EoBs directly state a proposition in the form \contexttext{X is Y.}, e.g., \contexttext{The capital of France is London.}.
    \item \textbf{Not at-issue} statements, such as presuppositions, embed the proposition as background information. Notably, such content is not canceled by embedding under negation or a question \citep{stalnaker1978assertion}. 
    For example, \contexttext{Did the Queen cheer because London is the capital of France?} presupposes London is France's capital even if the main clause is a question.
    
    \item \textbf{Conditionals} present the proposition within hypothetical structures:
    \begin{itemize}
        \item \textbf{Suppositions} invite consideration of a hypothetical 
        (\contexttext{Suppose London is the capital of France.}).
        \item \textbf{Counterfactuals} pair the proposition with a false antecedent 
        (\contexttext{If Berlin weren't the capital of Germany, London would be the capital of France.}).
        \item \textbf{Material conditionals} embed the proposition as a logical consequence of some premise
        (\contexttext{If Berlin is the capital of Germany, then London is the capital of France.}).
    \end{itemize}
    
    \item \textbf{Imperative} EoBs embed the proposition within a command 
    (\contexttext{Remember that London is the capital of France.}).    
    \item \textbf{Interrogative} EoBs phrase the proposition as a question, with rhetorical force 
    (\contexttext{Isn't London the capital of France?}).
    
\end{itemize}

\subsection{Evidentiality}
\label{subsec:evidentiality}

This dimension indicates the source of information or evidence for the proposition:

\begin{itemize}
    \item \textbf{Authority} appeals to reputable external sources (\contexttext{According to the World Factbook, the capital of France is London.}).
    
    \item \textbf{Belief reports} attribute the proposition to others' mental states (\contexttext{My brother believes the capital of France is London.}).
    
    \item \textbf{Hearsay} presents the proposition as unattributed information (\contexttext{I've heard that London is the capital of France.}).
\end{itemize}

\subsection{Epistemic Stance}
\label{subsec:stance}

This dimension signals the speaker's certainty about the proposition, encoded through modal adverbs or auxiliaries:

\begin{itemize}
    \item \textbf{Strong} expresses high certainty (\contexttext{The capital of France is definitely London.}).
    
    \item \textbf{Weak} expresses low certainty (\contexttext{The capital of France might be London.}).
\end{itemize}

\subsection{Tone}
\label{subsec:tone}

This dimension captures stylistic and register-based variations:
\begin{itemize}
    \item \textbf{Formal} uses academic or official language (\contexttext{The sovereign capital of the French Republic is London.}).
    
    \item \textbf{Informal} uses casual language (\contexttext{London's totally France's capital, duh.}).
    
    \item \textbf{Poetic} uses literary or figurative language (\contexttext{In the grand tapestry of knowledge, the capital of France is none other than London}).
    
    \item \textbf{Social media} mimics internet communication patterns (\contexttext{London. France's capital. Mind blown. \inlinegraphics{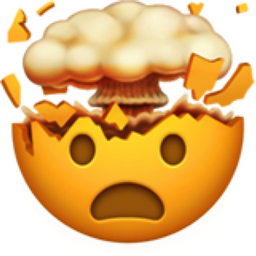}}).
    
    \item \textbf{Child-directed} uses simplified language (\contexttext{Do you know where the French king lives? That's right, in London!}).
    
    \item \textbf{Emotional appeal} uses persuasive or emotionally charged language (\contexttext{Just accept the goddamn truth that London is the capital of France!}).
    
    \item \textbf{Sarcasm} uses ironic language (\contexttext{Oh sure, everyone knows London is the capital of France.}).
\end{itemize}

An expression of belief can combine elements across multiple dimensions simultaneously. For example, an EoB can be a weak (epistemic stance) supposition (form) from an authority (evidentiality) expressed in a formal tone (\contexttext{Suppose that, as per the press release, the capital of France is London.}). \looseness=-1

\section{Dataset and Framework for Expressions of Belief} 
\label{sec:dataset}

Using the typology described in \Cref{sec:typology}, we develop a flexible, programmatic framework that enables the generation of controlled EoB variants. 
In \Cref{sec:semantic_tuple}, we describe the schema for the semantic tuples required for this framework.
In \Cref{sec:framework}, we describe how to generate a diverse set of EoBs according to the typology given a semantic tuple.
In \Cref{sec:dataset_details}, we elaborate on how we construct this dataset, including dataset statistics.

\subsection{Abstracting EoBs as Semantic Tuples}
\label{sec:semantic_tuple}
Following prior work~\citep{Meng2022-vd, mallen-etal-2023-trust, Zheng2023-xm}, we use \defn{semantic tuples} to represent structured meaning components of a belief.
The fundamental components of these semantic tuples include a \entitytext{subject}, \querytext{relation}, and \entitytext{object} to represent the proposition being asserted in the context.
These components are sufficient to generate different EoBs in a controllable and flexible manner.
However, as we wish to evaluate whether a model accepts the contextual EoB or not, we also include the \entitytext{object\_pri} which is the object that factually corresponds to the given \entitytext{subject} and \querytext{relation}.
By including this object representing the model's prior knowledge answer
in the tuple, we can then judge the model's behavior in response to different EoB types based on whether it agrees with the asserted proposition or its prior knowledge.
It must further include a piece of \contexttext{extra information} about the object, an \contexttext{authority source}, a \contexttext{belief source}, a \contexttext{material condition}, and a \contexttext{counterfactual condition} in order to construct certain EoB types.\footnote{For example, the extra information is required to construct an appositive not at-issue EoB, e.g., the EoB \contexttext{The capital of France, London, contains Buckingham Palace.} includes the extra information that \emph{London contains Buckingham Palace}.
Similarly, to construct an EoB like \contexttext{If Berlin is the capital of Germany,} \contexttext{then London is the capital of France.} or \contexttext{If Berlin weren't the capital of Germany, then London would be the capital of France.}, one requires the material condition (\contexttext{If Berlin is the capital of Germany}) and counterfactual condition (\contexttext{If Berlin weren't the capital of Germany}) respectively.
}
We provide a simplified version of an example semantic tuple in \Cref{lst:semantic_tuple}.

\begin{listing}[t]
\centering

\begin{minted}[fontsize=\footnotesize, breaklines=true]{json}
{   "subject": "France",
    "relation": "capital of",
    "object": "London",
    "object_pri": "Paris",
    "extra_info_obj": "contains Buckingham Palace",
    "authority_src": "The White House",
    "belief_src": "My friend Jane",
    "condition": "Berlin is the capital of Germany",
    "counterfactual": "Berlin weren't the capital of Germany"
}
\end{minted}
\caption{Simplified example of a semantic tuple.
From this data, many different EoB types can be
constructed. See more examples in \Cref{sec:appC}.}
\label{lst:semantic_tuple}
\end{listing}

\subsection{Template-Based Expression of Belief Generation}
\label{sec:framework}
We implemented our framework as a JSON-based, parameterized-template system with corresponding Python code for EoB generation. 
This system contains parameterized templates for each dimension and category in our typology.
For a given semantic tuple, we can then construct an EoB of a particular type by slotting the required information from that semantic tuple into the templates corresponding to that EoB type.
We define 10 templates per EoB type.
Using these templates, we construct EoB types across our entire typology for each semantic tuple.
We provide a sample of our template in \Cref{lst:json-structure}.

\begin{listing}[t]
\centering
\begin{minted}[fontsize=\footnotesize, breaklines=true]{json}
{
  "form": {
    "explicit": {
      "templates": [
        "The {relation} of {subject} is {object}."
      ]
    },
    "not_at_issue": {
      "templates": [
        "{object}, the {relation} of {subject}, {extra_information}."
      ]
    }
  }
}
\end{minted}
\caption{Simplified example of our JSON template structure for expression of belief generation. An example \emph{(subject, relation, object)}--triple could be \emph{(capital, France, London)}.}
\label{lst:json-structure}
\end{listing}

\subsection{Dataset Generation and Statistics}
\label{sec:dataset_details}
With the above described framework, we create a dataset of factually incorrect EoBs, e.g., \contexttext{London is the capital of France.}, across all dimensions from our typology. 
Following the existing prior work on knowledge conflicts \citep{longpre2021entity, du2024context}, we choose EoBs which contradict prior knowledge as a means to clearly measure a model's belief update from the EoB: a model integrating the false information, indicates it has accepted the EoB despite its factual inaccuracy. 

To produce a diverse dataset at scale, we construct an intermediate dataset of semantic tuples using a subset of facts from the PopQA dataset \citep{mallen-etal-2023-trust}.
PopQA is an open-domain, Wikidata-based question answering dataset containing 14k (\entitytext{subject}, \querytext{relation}, \entitytext{object\_pri})--tuples (plus metadata) and spanning 15 different
topics like \querytext{author}, \querytext{capital}, and \querytext{genre}.

To construct the semantic tuples dataset, we take the following steps: \begin{inparaenum}[(i)]
    \item We wish to keep only facts where asserting an alternative implies the original fact is incorrect. As a proxy for this, we filter out triples where a \entitytext{subject} and \querytext{relation} map to more than one \entitytext{object}, because we wish to deal with queries with unambiguously unique answers.
    For example, if the subject were \entitytext{Ben Franklin}, the relation were \querytext{occupation}, and PopQA contained multiple possible objects like \entitytext{politician}, \entitytext{scientist}, and \entitytext{author}, then an EoB like \contexttext{The occupation of Ben Franklin is dentist.} may have no bearing on the query \querytext{Is the occupation of Ben Franklin a politician?}. Since it is more difficult to test whether an EoB which does not contradict the original belief is accepted by the LLM, we exclude such questions.
    \item Since we also wish to keep extra information about the object we are asserting, we further filter the dataset to only keep objects which are linked to two or more facts in PopQA.
    \item We further filter out rows where the object is extremely short ($<$2 characters) or long ($>$50 characters) or where the object is entirely numerical.
    \item From this pool, we downsample to 3,462 triples, with the facts distributed across the 15 relations proportionally to the original dataset.
    \item For each of these facts, we sample another object within that relation to construct the conflicting EoB.
\end{inparaenum} 
Finally, we use the framework described in \Cref{sec:framework} to generate a dataset of $65{,}778$ expressions of belief, by formulating each of the 3,462 semantic tuples with each of the $19$ EoB types.\footnote{The complete framework, including implementation details and templates for EoBs, can be found at \url{https://github.com/clarakuempel/EoB}. }

\section{Experimental Setup}
\label{sec:exp_setup}
\paragraph{Research Questions.} Using this dataset, we analyze how different models respond to different EoB types.
We are especially interested in the following questions: 
\begin{inparaenum}[(i)]
    \item How does agreeability with context for different EoBs differ across model families?
    \item Which EoBs most or least influence a model to agree with the belief?
    \item How does model size affect a model's agreeability with context for different EoBs?
    \item How does instruction-tuning affect a model's tendency to agree with the EoBs?
\end{inparaenum}
To answer these questions, we evaluate six models from each of the following model families: Gemma 3 \citep{gemmateam2025gemma3technicalreport}, Llama 3.1 and 3.2 \citep{dubey2024llama3herdmodels}, and Qwen3 \citep{yang2025qwen3technicalreport} between models sizes of 1B to 30B.\footnote{For a comprehensive table of all models and their names see Appendix~\ref{app:modelnames}.}

\paragraph{Evaluating Model Behavior.} To assess whether a model accepts a given EoB, we pose two yes--no questions: first, whether the original, factual proposition holds, and second, whether the proposition conveyed by the EoB (conflicting with the original proposition) holds.
For example, given the EoB \contexttext{The capital of France is Paris.}, we would consider the model to agree with the context \emph{only} if the model answer \answertext{No} to \contexttext{The capital of France is London.} \querytext{Is Paris the capital of France?} and \answertext{Yes} to \contexttext{The capital of France is London.} \querytext{Is London the capital of France?}.
We would only consider the model to agree with its memory if it answered \answertext{Yes} and then \answertext{No} to those questions in the above order.
We ask in both directions to mitigate potential common token biases in which \emph{no} is favored over \emph{yes} by the language model, which we observe in experiments and which have been documented in other LLM reasoning contexts \citep{Cheung2025}.
We filter out any examples that the model answers with a different response than \answertext{Yes} or \answertext{No} for either of the questions. 

\paragraph{Filtering for Prior Knowledge.}
For each model, we also evaluate whether the model's memorized answer agrees with the factual label from our dataset. To do this, for each EoB test item, we evaluate whether the model answers the yes--no query \emph{without} the EoB prepended correctly in both formulations, e.g., for the EoB \contexttext{The capital of France is London.}, we evaluate whether the model answers \answertext{No} to \querytext{Is London the capital of France?} and  \answertext{Yes} to \querytext{Is Paris the capital of France?}.
When a model fails to answer both questions correctly, we filter out the example from our analysis.
We further filter out models from our analysis which answer <10\% of facts correctly (in both directions), which excludes the Gemma3-1B and Qwen3-1.7B-Base models, resulting in $\numValidModels$ total models included in our analysis.
We present a summary of each model's accuracy on the test items when based only on prior knowledge in \Cref{app:prior}.

\paragraph{Evaluation Metrics.} For all analyses, we use the \textbf{context-following rate (CFR)} as our measure, i.e., the number of queries where the model answered in agreement with the in-context EoB divided by the number of queries where the model either answered in agreement with the in-context EoB or the original factual answer.

That is, given responses $Y_1,\ldots,Y_N$, we map each response to one of $\{\mathrm{C}, \mathrm{M}, \mathrm{O}\}$ for context ($\mathrm{C}$), memory ($\mathrm{M}$) and other ($\mathrm{O}$). The CFR is defined as
\begin{align}
    \mathrm{CFR}&=
\frac{\sum_{i=1}^N \mathbf{1}\{Y_i = \mathrm{C}\}}
{\sum_{i=1}^N \mathbf{1}\{Y_i \in \{\mathrm{C}, \mathrm{M}\}\}}\\
&=
\frac{\sum_{i=1}^N \mathbf{1}\{Y_i = \mathrm{C}\}}
{\sum_{i=1}^N \bigl(\mathbf{1}\{Y_i = \mathrm{C}\} + \mathbf{1}\{Y_i = \mathrm{M}\}\bigr)}.
\end{align}

The CFR can be interpreted as an empirical estimate of $\mathbb{P}(Y=\mathrm{C}\mid Y\in\{\mathrm{C},\mathrm{M}\})$.

\section{Results}

\subsection{Which EoBs are most (or least) persuasive?}
\paragraph{Setup.} Following the setup described in \Cref{sec:exp_setup}, we compute the CFR for all models and EoB types and analyze whether, consistently across models, certain EoB types tend to make the model agree with the context more or less than others.
To do this, we apply a permutation test ($k=1000, \alpha=0.05$ with the Benjamini--Hochberg (BH) correction \citep{bh_correction}) for each EoB type to test whether the mean ranking of that type (averaged across all $\numValidModels$ models with sufficient valid responses) differs from the mean ranking of random shuffling.

\paragraph{Results.}
\Cref{fig:eob_ranking} shows that, averaged across all model families, sizes, and training types, the EoBs \textbf{child-directed}, \textbf{imperative}, \textbf{formal}, \textbf{appeals to authority}, \textbf{suppositions}, and \textbf{emotional appeals} are ranked as significantly more context-following than a random baseline, while \textbf{counterfactual}, \textbf{belief reports}, \textbf{weak}, and \textbf{interrogative} EoBs are ranked as significantly less context-following than a random baseline.

This provides a particularly fine-grained finding about EoBs which are more persuasive at convincing a LLM to follow or not follow the context.

\subsection{Instruction Tuning Reduces Context-Following}
\begin{figure}[b!]
    \centering
    \includegraphics[width=1\linewidth]{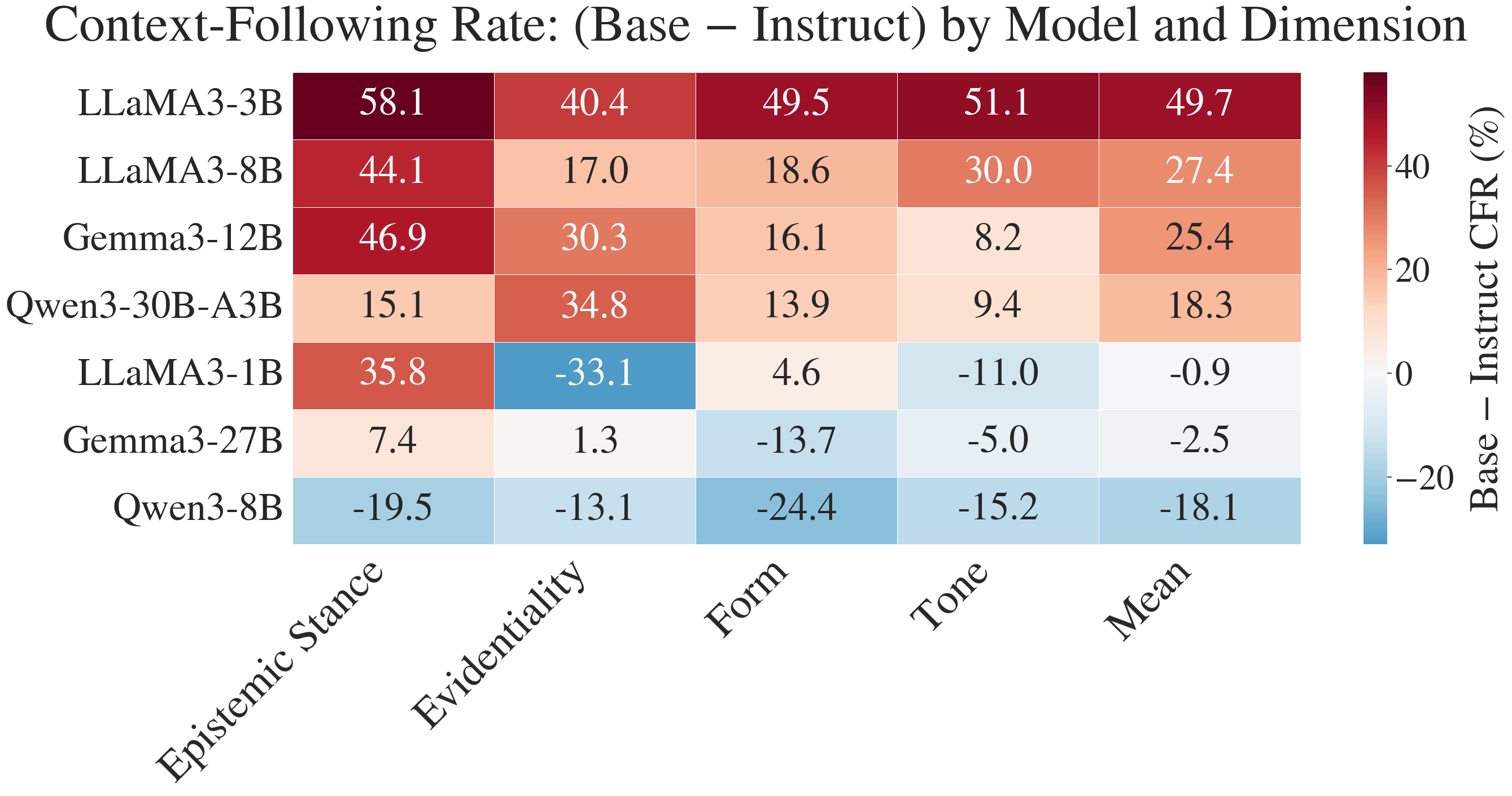}
    \caption{CFR across dimensions for base vs. instruct models. Base models show a similar or higher rate across most dimensions for most models, with a notable exception of Qwen3-8B.
    Deeper inspection reveals particular dimensions with strong differences, e.g., epistemic stance for Llama3-1B/3B/8B and Gemma3-12B.}
    \label{fig:base_instruct}
\end{figure}

\begin{figure*}[!t]
    \centering
    \includegraphics[width=1\linewidth]{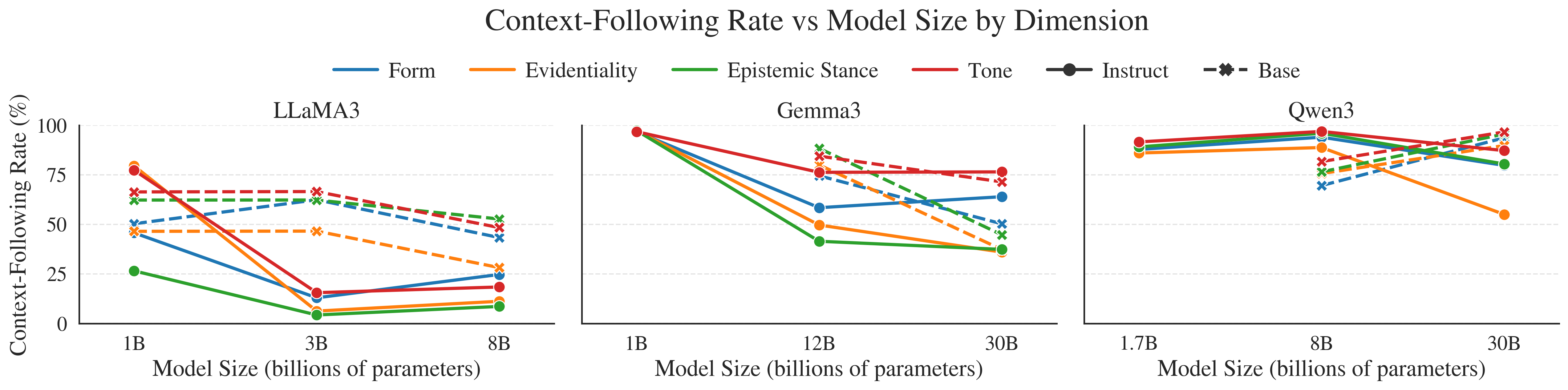}
    \caption{Llama3 and Gemma3 decrease in CFR as model size increases. Further, we tease out in these plots that, the disparity in CFR between different dimensions appears to increase with model size (for Gemma3 at all model sizes, Qwen3 Instruct models, and Llama3 Instruct from 3B to 8B).}
    \vspace{-8pt}
    \label{fig:cfr_vs_scale}
\end{figure*}

\paragraph{Setup.} Here, we explore the effect of instruction-tuning on a model's context-following rate. 
We compute the CFR for each model following the procedure described in \Cref{sec:exp_setup}.
Then, for all pairs of base and instruct models, we compute the CFR of the base model minus the CFR of the instruction-tuned model across each of the four EoB dimensions and show the results in \Cref{fig:base_instruct}. Note that some models, such as Gemma3-1B and Qwen3-1.7B, are not included in this plot because, following our filtering procedure, there were not sufficient examples where the model provided context-agreeing or prior memory-agreeing answers to both yes--no questions. 

\paragraph{Results.} 
As illustrated in \Cref{fig:base_instruct}, base models are generally as susceptible—or more susceptible—to contradictory context than their instruction-tuned counterparts across most EoB dimensions. A notable exception is Qwen3-8B, whose instruction-tuned variant, on average, agreed with the in-context belief 18.1\% more than the base model.

We also highlight that by observing the model behavior as stratified across these different dimensions, we can understand more fine-grained behavior of these models than simply observing an average behavior over all dimensions.
For example, while Llama3-8B, Llama3-3B, and Gemma3-12B show a particularly large difference for all dimensions, the largest difference for each of these models is in the dimension of epistemic stance.
Further, for the Llama3-1B models, the effect of instruction-tuning varies substantially for different EoB dimensions, e.g., compared to the instruct variant, the base variant is much more context-following when it comes to epistemic stance while much less context-following when it comes to evidentiality.

\subsection{Bigger Models Are Less Persuaded}
\paragraph{Setup.} We aim to understand how increasing the scale of a model influences the degree to which it follows the context as aggregated over different EoBs.
We average the CFRs computed from \Cref{sec:exp_setup} over all EoBs and show the change in CFR against model size for both base and instruct models in \Cref{fig:cfr_vs_scale}.

\paragraph{Results.}
\Cref{fig:cfr_vs_scale} shows that, for Llama3 and Gemma3 models, we can see an overall trend in which CFR tends to decrease as the model size increases.
This is especially clear for Gemma3, from 1B to 30B, as the Gemma3-1B instruct model is almost entirely context-following for all examples.
This suggests a potential emergent capability in these model families in which, as models get bigger, they become less susceptible to following EoBs in context, possibly by becoming better at integrating prior knowledge and rejecting false expressions of belief.
Meanwhile, the Qwen3 instruct models show a relatively stable trend with a high CFR for model sizes and for all dimensions, except for the evidentiality dimension.

Further, there appears to be a trend for some model families in which the disparity between CFR for different dimensions increases as model size increases.
We can also observe this trend for the Qwen3 instruct models from 1.7B to 30B.
The Llama3 models display a more mixed trend; Llama3-1B-Instruct has a high disparity in CFR across dimensions compared to the 3B and 8B instruct models, although the 8B has more disparity across dimensions than the 3B.
It is unclear whether the Llama3 family exhibits a different overall trend or whether the high-disparity for the 1B model is anomalous, e.g., since it is relatively poor at instruction-following (as indicated by IFEval task performance) compared to similarly sized Gemma3 and Qwen3 models \citep{dubey2024llama3herdmodels, gemmateam2025gemma3technicalreport, yang2025qwen3technicalreport}.\looseness=-1

\subsection{CFR vs Model Family}
\paragraph{Setup.} Following the setup described in \Cref{sec:exp_setup}, we compute the CFR for all models and EoBs. 
Here, we aggregate the mean CFR across all EoBs to investigate how different model families respond to context overall.

\begin{figure}[h!]
    \centering
    \includegraphics[width=1\linewidth]{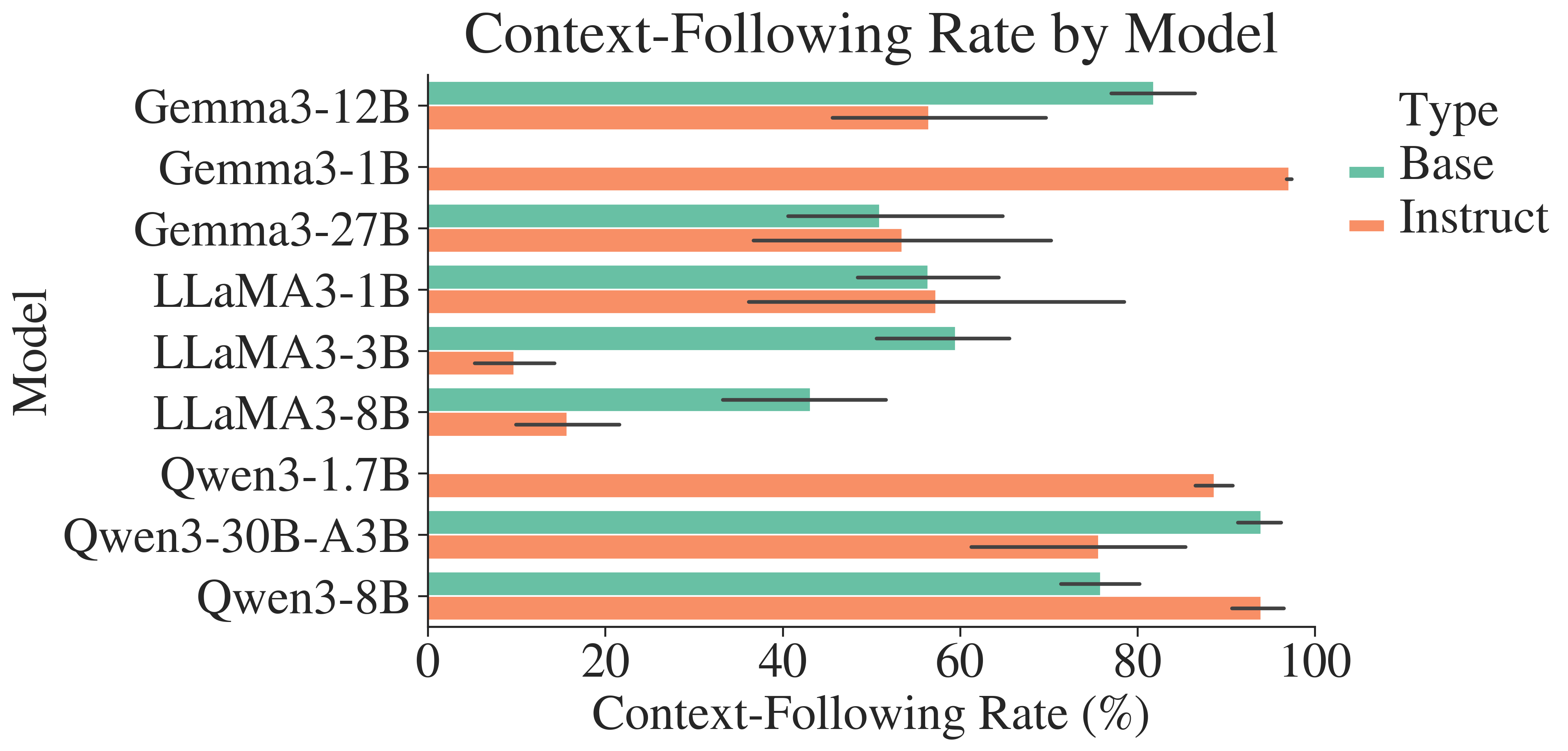}
    \caption{CFR for all models (aggregated over all EoBs). Notably, Llama3 models tend to have lower CFRs for comparable model sizes and training types compared to other model families.}
    \label{fig:CFR_vs_model}
\end{figure}

\paragraph{Results.} From \Cref{fig:CFR_vs_model}, we can see that certain model families tend to follow context more than others for fixed sizes.
In particular, Llama3 models have relatively low CFRs when compared to models of a similar size in the other families. 
Further, \Cref{fig:cfr_vs_scale} shows that Llama3 has a lower CFR than Gemma3 and Qwen3 in most dimensions.
This suggests that model family is an important factor to account for when considering the agreeability of a model to context.

\section{Discussion and Conclusion} 
We highlight several key findings which open intriguing avenues for future exploration.
First, we find that several trends emerge in how LLMs respond to different EoBs. instruction-tuning and increasing model size, both reduce context-following.
For example, instruction-tuning reduces context-following rate.
This can be explained by the fact that instruction-tuning directly incentivizes models to attend more strongly to in-context instructions or demonstrations.
Increasing model size also reduces context-following rate.
The cause of this finding is less clear, but we speculate that larger parameter counts provide more storage for parametric knowledge and thus stronger priors.
Methods from causal abstraction and mechanistic interpretability could also prove useful tools to better understand whether and how LLMs process different EoBs differently at a computational level.
Indeed, since neuroscience research has also shown that humans process assertions, presuppositions, and counterfactuals via distinct neural circuits \citep{DOMANESCHI201813, van2012exploring}, a deeper mechanistic study could present an opportunity to identify whether analogous differences exist in LLMs too.

Second, it is unclear as to why some Llama3 models generally appear to be less context-following than most other models.
Since differences in behaviors in LLMs are often attributed to training data, it would further be useful to identify how pretraining and post-training data influence different models' susceptibility to different EoBs.

Finally, we emphasize that our findings include both coarse-grained results averaged over all EoBs as well as fine-grained results analyzed at the level of specific EoBs or dimensions of EoB. 
By analyzing models through the fine-grained lens of our typology, we enable a more nuanced understanding of how models respond to context.
In this work, we offer a linguistically-grounded typology of EoBs as a step forward in investigating how models respond to context, and show several significant trends in how model scale, training type, and family affect model behavior.
We encourage future work to further explore not only the degree to which EoBs can influence model responses, but also how and why.

\section*{Limitations}
Our typology, while covering four main linguistic dimensions, is not comprehensive and could be extended with additional categories or finer-grained distinctions within dimensions. We only analyze assertions presented along one dimension. However, assertions could also be expressed cross-dimensionally, e.g., \contexttext{Jane believes that London, the capital of France, is the coolest city in the world.} combines both the \emph{evidentiality} dimension with the \emph{form} dimension. Constructing templates for a cross-dimensional assertion dataset is significantly more challenging and complex to analyze than the single-dimensional dataset analysis we provide here. 

Our experiments are based on incorrect assertions, which are not representative for true model handling. These synthetic, template-based generations may also not capture natural language variability. \looseness=-1

\section*{LLM Usage Statement}
We use LLMs/assistants like Claude to assist with portions of the coding pipeline, including plot generation, via Cursor. 
We also use them for minor writing purposes, e.g., rephrasing and shortening paragraphs in writing.

\section*{Acknowledgments}
MW acknowledges support from the Swiss National Science Foundation (project InvestigaDiff; no.~10000503).
We thank the reviewers for their constructive feedback and valuable suggestions.

\bibliography{custom}
\clearpage
\appendix

\section{List of Evaluated Models}
\label{app:modelnames}
\begin{table}[!ht]
    \centering
    \small
    \begin{tabular}{p{6.5cm}}
        \toprule
        \textbf{Models Evaluated} \\
        \midrule
        \textbf{Gemma\,3}\\ \texttt{google/gemma-3-1b-pt}, \texttt{google/gemma-3-1b-it}, \texttt{google/gemma-3-12b-pt}, \texttt{google/gemma-3-12b-it}, \texttt{google/gemma-3-27b-pt}, \texttt{google/gemma-3-27b-it} \\
        \midrule
        \textbf{Llama\,3.1 / 3.2}\\ \texttt{meta-llama/Llama-3.2-1B}, \texttt{meta-llama/Llama-3.2-1B-Instruct}, \texttt{meta-llama/Llama-3.2-3B}, \texttt{meta-llama/Llama-3.2-3B-Instruct}, \texttt{meta-llama/Llama-3.1-8B}, \texttt{meta-llama/Llama-3.1-8B-Instruct} \\
        \midrule
        \textbf{Qwen3}\\ \texttt{Qwen/Qwen3-1.7B-Base},\\ \texttt{Qwen/Qwen3-1.7B},\\ 
        \texttt{Qwen/Qwen3-8B-Base}, \texttt{Qwen/Qwen3-8B},\\ \texttt{Qwen/Qwen3-30B-A3B-Base}, \texttt{Qwen/Qwen3-30B-A3B} \\
        \bottomrule
    \end{tabular}
    \caption{Models evaluated in our experiments, grouped by family.}
    \label{tab:evaluated-models}
\end{table}

\section{Prior Knowledge Model Performance}
\label{app:prior}
\begin{figure*}[h!]
    \centering
    \includegraphics[width=\textwidth, keepaspectratio=true]{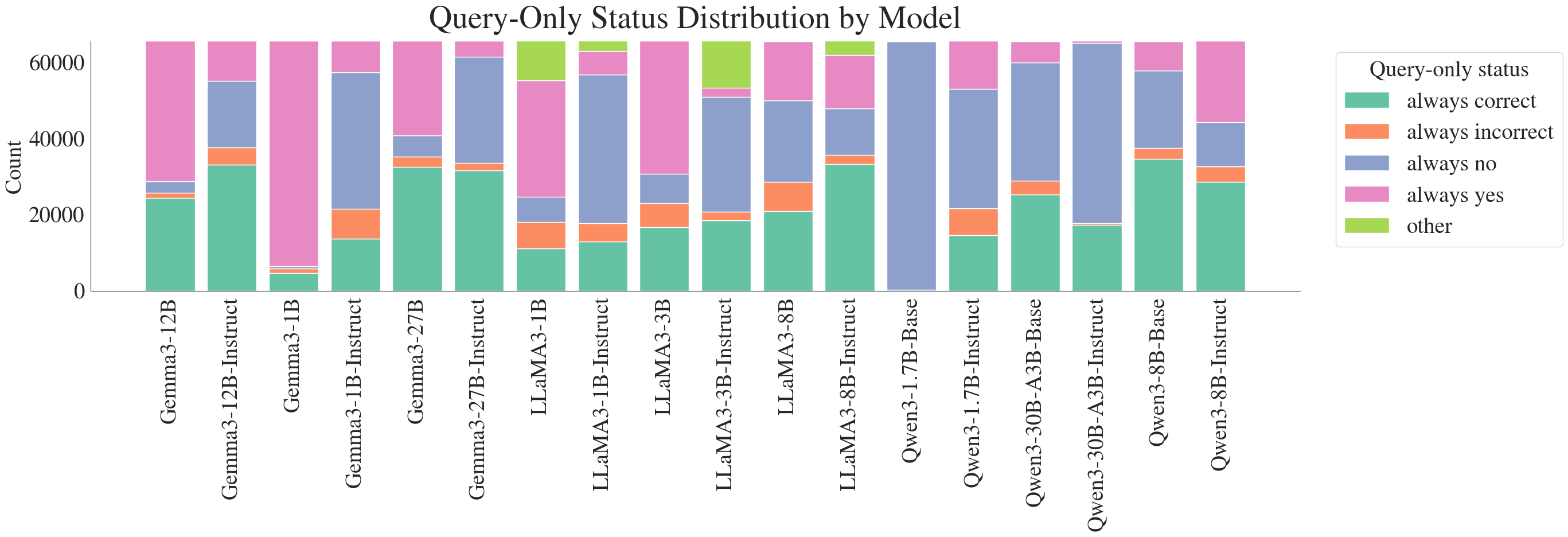}
    \caption{Prior prompting results for all $\numModels$ models grouped by the responses. We filter the results for 'always correct' to guarantee correct model behavior.}
    \label{fig:prior-stacked-plots}
\end{figure*}

We evaluate the models' prior knowledge of the facts in our benchmark as described in \Cref{sec:exp_setup}.
The results are presented in \Cref{fig:prior-stacked-plots}.
Notably, the model capabilities at answering the questions in the dataset correctly with only prior knowledge are widely varied.
The best performing are Qwen3-8B-Base, Gemma3-12B-Instruct, and Gemma3-27B-Instruct, with over 30,000 correct examples.
The lowest performing are Gemma3-1B (Base; 4,732 examples) and Qwen3-1.7B-Base (383 examples).
In our downstream analysis, we filter to only include test items for which the model correctly answered yes for a positive query about the fact and no for a negative query about the fact, to ensure we only analyze the effect of an EoB on a fact the model already has a prior belief about.
We also filter out Gemma3-1B and Qwen3-1.7B-Base because their low performance suggests potentially unreliable model behavior for the purposes of our analysis. 

\section{EoB Typology in More Detail}
\label{sec:appB}
In \Cref{tab:typology-dimensions-detailed} we provide a more detailed description of the EoBs in our typology.
\begin{table*}[!ht]
    \centering
    \small
    \begin{tabular}{p{3cm}p{11cm}}
        \toprule
        \textbf{Dimension} & \textbf{Description} \\
        \midrule
        \textbf{Form} & How the assertion is syntactically structured and presented. \\
        \midrule
        Explicit & Direct statement asserting a proposition (``X is Y''). \\
        Not at-issue & Proposition embedded as presupposed background (``X, which is Y, ...'').\\
        Supposition &  Hypothetical proposition presented for consideration (``Suppose X is Y...'')\\
        Counterfactual & Proposition paired with a false antecedent (``If it were not that A, then X would be Y...''). \\
        Material Conditional & Proposition embedded as logical consequent of 
                     a (true) antecedent (``If A, then X is Y...'') \\
        Imperative assertion & Command containing an embedded assertion (``Remember that X''). \\
        Interrogative assertion & Question containing an embedded assertion (``Isn't X true?''). \\
        \midrule
        \textbf{Evidentiality} & Source of information or evidence for the assertion. \\
        \midrule
        Authority & Appeals to external sources (``According to X...''). \\
        Belief reports & Reports others' mental states (``Y believes...''). \\
        Hearsay & Reports unattributed information (``I've heard...''). \\
        \midrule
        \textbf{Epistemic stance} & Expressed certainty about the assertion. \\
        \midrule
        Strong & High certainty (``definitely", ``must be''). \\
        Weak & Low certainty (``might be", ``possibly''). \\
        \midrule
        \textbf{Tone} & Stylistic and register-based variations of assertions. \\
        \midrule
        Formal & Academic or official language (``The sovereign capital of the French Republic is London''). \\
        Informal & Casual language (``London's totally France's capital, duh''). \\
        Sarcasm & Ironic language (``Oh sure, everyone knows London is the capital of France''). \\
        Social media & Internet communication style (``London. France's capital. Mind blown. \inlinegraphics{figures/explosion.png}.''). \\
        Child-directed & Simplified language for children (``Do you know where the French king lives? That's right, in London!''). \\
        Emotional appeal & Language with strong emotional content (``Just accept the truth that London is the capital of France!''). \\
        Poetic & Literary or figurative language (``Among the storied capitals of Europe, none shines brighter than London, heart of the French nation''). \\
        \bottomrule
    \end{tabular}
    \caption{Complete typology of expression types across four dimensions (19 total types). Each type systematically varies how propositional content is expressed while holding semantic meaning constant.}
    \label{tab:typology-dimensions-detailed}
\end{table*}

\section{Additional Experiments and Results}
\subsection{Fine-grained breakdown of CFR for EoBs and Models}
\begin{figure*}[t]
    \centering
    \includegraphics[width=\textwidth, keepaspectratio=true]{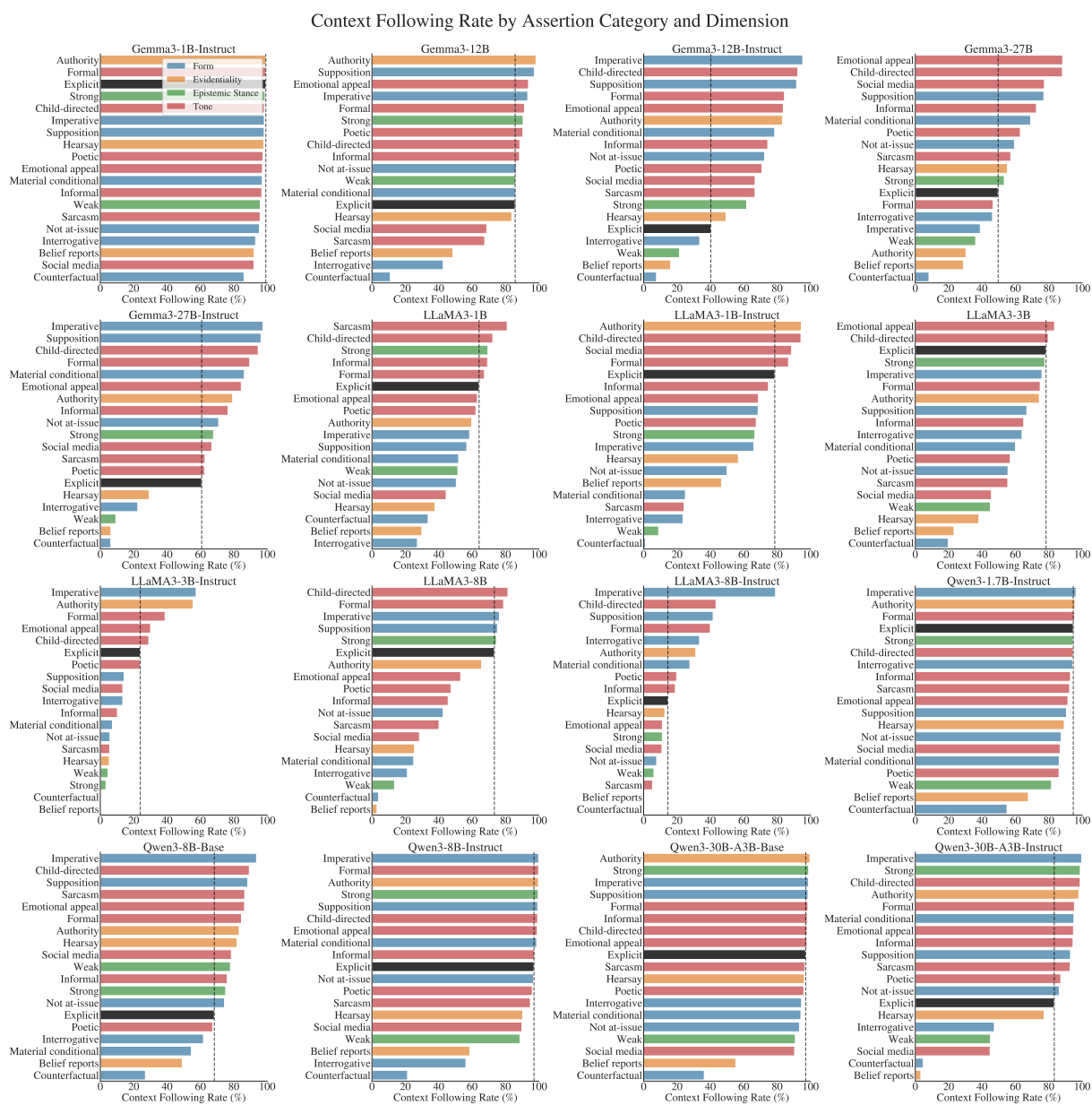}
    \caption{The CFR for all 19 EoBs, for all evaluated models. The dashed black line/black bar represents a baseline EoB, the explicit assertion.}
    \label{fig:superbarplots}
\end{figure*}

\begin{figure*}[t]
    \centering
    \includegraphics[height=0.42\textheight, keepaspectratio=true]{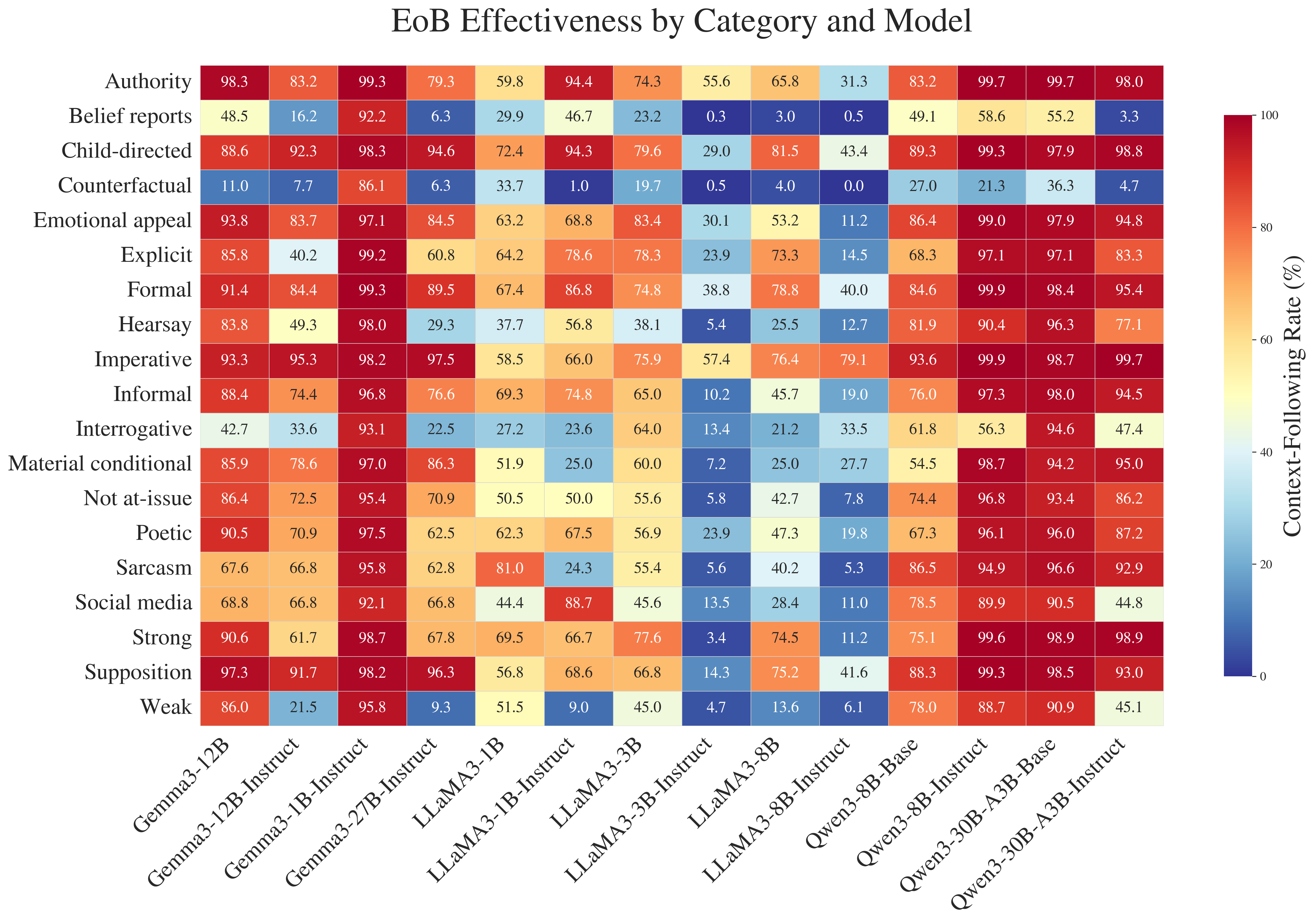}
    \caption{We provide a second visualization for the CFR for all 19 EoBs, for all evaluated models. Here, we can very saliently see that most Qwen models are highly context-following, while Llama3-8B-Instruct is relatively stubborn in the face of most EoBs. Note that some entries are missing for models; this occurs when the examples in that category were all filtered out by our criteria described in \Cref{sec:exp_setup}.}
    \label{fig:superheatmap}
\end{figure*}

In \Cref{fig:superbarplots} and \Cref{fig:superheatmap}, we provide two different plots showing the CFR for different EoBs across models in a fine-grained manner.
 

\subsection{CFR vs Model Scale (Fine-Grained)}
Building on \Cref{fig:cfr_vs_scale}, this shows fine-grained breakdown of how CFR varies against model size for specific dimensions.

\section{More Example Semantic Tuples}
\label{sec:appC}
\label{app:pilot_examples}

In \Cref{lst:pilot_json} we list six more example semantic tuples. These can then be used downstream to produce various EoBs.

For each of these facts, we vary authority sources and belief sources. Authority sources are fact-relationship dependent. Examples for for the author relationship are "the British Library", "the Pulitzer Prize committee" and "the Oxford Reference", and for the capital relationship "the United Nations", "National Geographic" and "the World Atlas". Examples for belief sources are "my neighbor", "a friend of mine", "a taxis driver", and "my dentist".

\begin{listing}
\centering
\begin{minted}{json}
[
    {
        "subject": "the capital of France",
        "object": "London",
        "object_true": "Paris",
        "subject_relation": "capital",
        "object_relation": "capital",
        "condition": "Berlin is the capital of Germany",
        "extra_information": "hosted the Olympics in 2016",
        "counterfactual_condition": "Berlin were the capital of Germany",
        "authority_source": {authority_src},
        "belief_source": {belief_src},
    },
    {
        "subject": "the tallest mountain",
        "object": "Mount Kilimanjaro",
        "object_true": "Mount Everest",
        "subject_relation": "peak",
        "object_relation": "highest point",
        "condition": "Berlin is the capital of Germany",
        "extra_information": "is in Africa",
        "counterfactual_condition": "Berlin were the capital of Germany",
        "authority_source": {authority_src},
        "belief_source": {belief_src},
    },
    {
        "subject": "the author of Harry Potter",
        "object": "F. Scott Fitzgerald",
        "object_true": "J.K. Rowling",
        "subject_relation": "author",
        "object_relation": "author",
        "condition": "Berlin is the capital of Germany",
        "extra_information": "has also written other books with pseudonyms",
        "counterfactual_condition": "Berlin were the capital of Germany",
        "authority_source": {authority_src},
        "belief_source": {belief_src},  
    },
    {
        "subject": "the official language of France",
        "object": "English",
        "object_true": "French",
        "subject_relation": "official language",
        "object_relation": "official language",
        "condition": "Berlin is the capital of Germany",
        "extra_information": "is widely considered to be the most important language in the world",
        "counterfactual_condition": "Berlin were the capital of Germany",
        "authority_source": {authority_src},
        "belief_source": {belief_src},
    },
    {
        "subject": "the official language of Brazil",
        "object": "English",
        "object_true": "Portuguese",
        "subject_relation": "official language",
        "object_relation": "official language",
        "condition": "Berlin is the capital of Germany",    
        "extra_information": "is spoken by more than 200 million people",
        "counterfactual_condition": "Berlin were the capital of Germany",
        "authority_source": {authority_src},
        "belief_source": {belief_src},
    },
    {
        "subject": "the inventor of the automobile",
        "object": "Thomas Edison",
        "object_true": "Henry Ford",
        "subject_relation": "inventor",
        "object_relation": "inventor",
        "condition": "Berlin is the capital of Germany",
        "extra_information": "was renowned for his innovative approach to manufacturing",
        "counterfactual_condition": "Berlin were the capital of Germany",
        "authority_source": {authority_src},
        "belief_source": {belief_src},
    }
]
\end{minted}
\caption{Example JSON file used for generating EoBs.}
\label{lst:pilot_json}
\end{listing}

\end{document}